%% file: main.tex
\documentclass{article}

     \usepackage[final]{neurips_rl2019}

\input{math_commands.tex}

\usepackage[utf8]{inputenc} % allow utf-8 input
\usepackage[T1]{fontenc}    % use 8-bit T1 fonts
\usepackage{hyperref}       % hyperlinks
\usepackage{url}            % simple URL typesetting
\usepackage{booktabs}       % professional-quality tables
\usepackage{amsfonts}       % blackboard math symbols
\usepackage{nicefrac}       % compact symbols for 1/2, etc.
\usepackage{microtype}      % microtypography

\usepackage{caption}
\usepackage{subcaption}
\usepackage{graphicx}

\usepackage[acronym]{glossaries}

\newacronym{mpn}{MPN}{Multimodal Policy Network}

\title{Imitation Learning of Robot Policies by Combining Language, Vision and Demonstration}

\author{
  Simon Stepputtis$^1$, Joseph Campbell$^1$, Mariano Phielipp$^2$, Chitta Baral$^1$, Heni Ben Amor$^1$ \\
  $^1$ School of Computing, Informatics, and Decision Systems Engineering, ASU \\
  \texttt{\{sstepput, jacampb1, chitta, hbenamor\}@asu.edu}\\
  $^2$ Intel AI Lab\\
  \texttt{mariano.j.phielipp@intel.com} \\
}

\begin{document}

\maketitle

\begin{abstract}
In this work we propose a novel end-to-end imitation learning approach which combines natural language, vision, and motion information to produce an abstract representation of a task, which in turn is used to synthesize specific motion controllers at run-time. This multimodal approach enables generalization to a wide variety of environmental conditions and allows an end-user to direct a robot policy through verbal communication. 
We empirically validate our approach with an extensive set of simulations and show that it achieves a high task success rate over a variety of conditions while remaining amenable to probabilistic interpretability.
\end{abstract}

\section{Introduction}
A significant challenge when designing robots to operate in the real world lies in the generation of control policies that can adapt to changing environments. Programming such policies is a labor and time-consuming process which requires substantial technical expertise. Imitation learning~\citep{schaal1999imitation}, is an appealing methodology that aims at overcoming this challenge -- instead of complex programming, the user only provides a set of demonstrations of the intended behavior. These demonstrations are consequently distilled into a robot control policy by learning appropriate parameter settings of the controller. Popular approaches to imitation, such as Dynamic Motor Primitives (DMPs)~\citep{ijspeert2013dynamical} or Gaussian Mixture Regression (GMR)~\citep{calinon2009robot} largely focus on motion as the sole input and output modality, i.e., joint angles, forces or positions. Critical semantic and visual information regarding the task, such as the appearance of the target object or the type of task performed, is not taken into account during training and reproduction. The result is often a limited  generalization capability which largely revolves around adaptation to changes in the object position. 
While imitation learning has been successfully applied to a wide range of tasks including table-tennis~\cite{mulling2013learning}, locomotion~\cite{chalodhorn2007learning}, and human-robot interaction~\cite{amor2014interaction} an important question is how to incorporate language and vision into a differentiable end-to-end system for complex robot control.

In this paper, we present an imitation learning approach that combines language, vision, and motion in order to synthesize natural language-conditioned control policies that have strong generalization capabilities while also capturing the semantics of the task.
We argue that such a multi-modal teaching approach enables robots to acquire complex policies that generalize to a wide variety of environmental conditions based on descriptions of the intended task. 
In turn, the network produces control parameters for a lower-level control policy that can be run on a robot to synthesize the corresponding motion. 
The hierarchical nature of our approach, i.e., a high-level policy generating the parameters of a lower-level policy, allows for generalization of the trained task to a variety of spatial, visual and contextual changes.

\paragraph{Problem Statement:} In order to outline our problem statement, we contrast our approach to Imitation learning~\citep{schaal1999imitation} which considers the problem of learning a policy $\mathbf{\pi}$ from a given set of demonstrations ${\cal D}=\{\mathbf{d}^0,.., \mathbf{d}^m\}$. Each demonstration spans a time horizon $T$ and contains information about the robot's states and actions, e.g., demonstrated sensor values and control inputs at each time step. Robot states at each time step within a demonstration are denoted by $\mathbf{x}_t$. 
In contrast to other imitation learning approaches, we assume that we have access to the raw camera images of the robot $\mI_t$ at teach time step, as well as access to a verbal description of the task in natural language. This description may provide critical information about the context, goals or objects involved in the task and is denoted as $\mathbf{s}$. Given this information, our overall objective is to learn a policy $\mathbf{\pi}$ which imitates the demonstrated behavior, while also capturing semantics and important visual features. After training, we can provide the policy $\mathbf{\pi}(\mathbf{s},\mI)$ with a different, new state of the robot and a new verbal description (instruction) as parameters. The policy will then generate the control signals needed to perform the task which takes the new visual input and semantic context int
o account. 

\section{Background}
A fundamental challenge in imitation learning is the extraction of policies that do not only cover the trained scenarios, but also generalize to a wide range of other situations.
A large body of literature has addressed the problem of learning robot motor skills by imitation~\citep{argall2009survey}, learning functional~\citep{ijspeert2013dynamical} or probabilistic~\citep{maeda2014learning} representations.
However, in most of these approaches, the state vector has to be carefully designed in order to ensure that all necessary information for adaptation is available.
Neural approaches to imitation learning~\citep{pomerleau1989alvinn} circumvent this problem by learning suitable feature representations from rich data sources for each task or for a sequence of tasks \citep{Burke2019, Hristov2019, Misra2018}. 
Many of these approaches assume that either a sufficiently large set of motion primitives is already available or that a taxonomy of the task is available, i.e., semantics and motions are not trained in conjunction.
The importance of maintaining this connection has been shown in \citet{Chang}, allowing the robot to adapt to untrained variations of the same task. To learn entirely new tasks, meta-learning aims at learning policy parameters that can quickly be fine-tuned to new tasks \citep{finn17a}. While very successful in dealing with visual and spatial information, these approaches do not incorporate any semantic or linguistic component into the learning process. Language has shown to successfully generate task descriptions in \citet{Arumugam2019} and several works have investigated the idea of combining natural language and imitation learning:~\citet{Nicolescu:2003:NMR:860575.860614, Gemignani:2015:TRP:2772879.2773262, Cederborg:2013:LMG:2712980.2713055, Mericli:2014:IAS:2617388.2617416, Sugita2005}. 
However, most approaches do not utilize the inherent connection between semantic task descriptions and low-level motions to train a model. 

Our work is most closely related to the framework introduced in ~\citet{Tellex2014}, which also focuses on the symbol grounding problem. More specifically, the work in ~\citet{Tellex2014} aims at mapping perceptual features in the external world to constituents in an expert-provided natural language instruction. 
Our work approaches the problem of generating dynamic robot policies by fundamentally combining language, vision, and motion control in to a single differentiable neural network that can learn the cross-modal relationships found in the data with minimal human feature engineering. 
Unlike previous work, our proposed model is capable of directly generating complex low-level control policies from language and vision that allow the robot to reassemble motions shown during training.

\section{Multimodal Policy Generation via Imitation}

\begin{figure}
    \centering
    \includegraphics[width=1.0\textwidth]{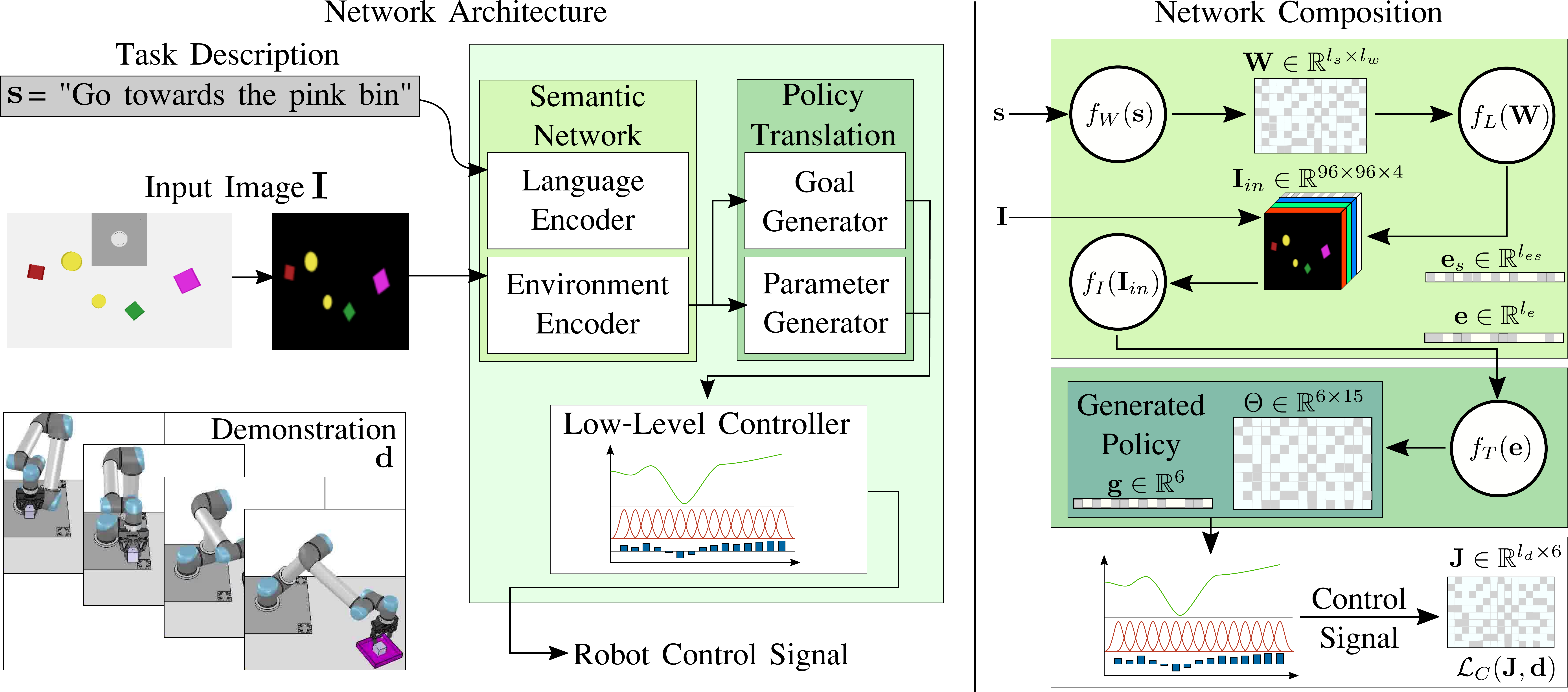}
    \caption{Network architecture overview. The network consists of two parts, a high-level semantic network and a low-level control network. Both networks are working seamlessly together and are utilized in an End-to-End fashion.}
    \label{fig:system}
\end{figure}\vspace{-1em}

We motivate our approach with a simple example: consider a binning task in which a 6 DOF robot has to drop an object into one of several differently shaped and colored bowls on a table.
To teach this task, the human demonstrator does not only provide a kinesthetic demonstration of the desired trajectory, but also a verbal command, e.g., \textit{``Move towards the blue bowl''} to the robot. 
In this example, the trajectory generation would have to be conditioned on the \emph{blue} bowl's position which, however, has to be extracted from visual sensing.
Our approach automatically detects and extracts these relationships between vision, language, and motion modalities in order to make best usage of contextual information for better generalization and disambiguation.

Figure~\ref{fig:system} (left) provides an overview of our method. Our goal is to train a deep neural network that can take as input a task description $\mathbf{s}$ and and image $\mI$ and consequently generates robot controls. In the remainder of this paper, we will refer to our network as the \gls{mpn}. Rather than immediately producing control signals, the \gls{mpn} will generate the parameters for a lower-level controller. This distinction allows us to build upon well-established control schemes in robotics and optimal control. In our specific case, we use the widely used Dynamic Motor Primitives~\citep{ijspeert2013dynamical} as a lower-level controller for control signal generation.   

In essence, our network can be divided into three parts. The first part, the semantic network, is used to create a task embedding $\ve$ from the input sentence $\vs$ and environment image $\mI$. In a first step, the sentence $\vs$ is tokenized and converted into a sentence matrix $\boldsymbol{W} \in \mathbb{R}^{l_s \times l_w} = f_W(\vs)$ by utilizing pre-trained Glove word embeddings \citep{pennington2014glove} where $l_s$ is the padded-fixed-size length of the sentence and $l_w$ is the size of the glove word vectors. 
To extract the relationships between the words, we use use multiple CNNs $\ve_s = f_L(\mW)$ with filter size $n \times l_w$ for varying $n$, representing different $n$-gram sizes~\citep{san}. The final representation is built by flattening the individual $n$-grams with max-pooling of size $(l_s - n_i + 1)\times l_w$ and concatenating the results before using a single perceptron to detect relationships between different $n$-grams. In order to combine the sentence embedding $\ve_s$ with the image, it is concatenated as a fourth channel to the input image $\mI$. The task embedding $\ve$ is produced with three blocks of convolutional layers, composed of two regular convolutions, followed by a residual convolution~\citep{resnet} each.

In the second part, the policy translation network is used to generate the task parameters $\Theta \in \mathcal{R}^{o \times b}$ and $\vg \in \mathcal{R}^{o}$ given a task embedding $\ve$ where $o$ is the number of output dimensions and $b$ the number of basis functions in the DMP:
\begin{equation}
    \boldsymbol{\Theta}, \vg = f_T(\ve) = f_G\left(\text{ReLU}\left(\mW_G \ve + \vb_G\right)\right), f_H \left(\text{ReLU}\left(\mW_G \ve + \vb_G\right)\right)
\end{equation}
where $f_G()$ and $f_H()$ are multilayer-perceptrons that use $\ve$ after being processed in a single perceptron with weight $\mW_G$ and bias $\vb_G$. These parameters are then used in the third part of the network, which is a DMP~\citep{schaal1999imitation}, allowing us leverage a large body of research regarding their behavior and stability, while also allowing other extensions of DMPs~\citep{amor2014interaction, paraschos2013probabilistic, khansari2011learning} to be incorporated to our framework.  

\section{Results}
\begin{figure}
    \centering
    \includegraphics[width=0.8\textwidth]{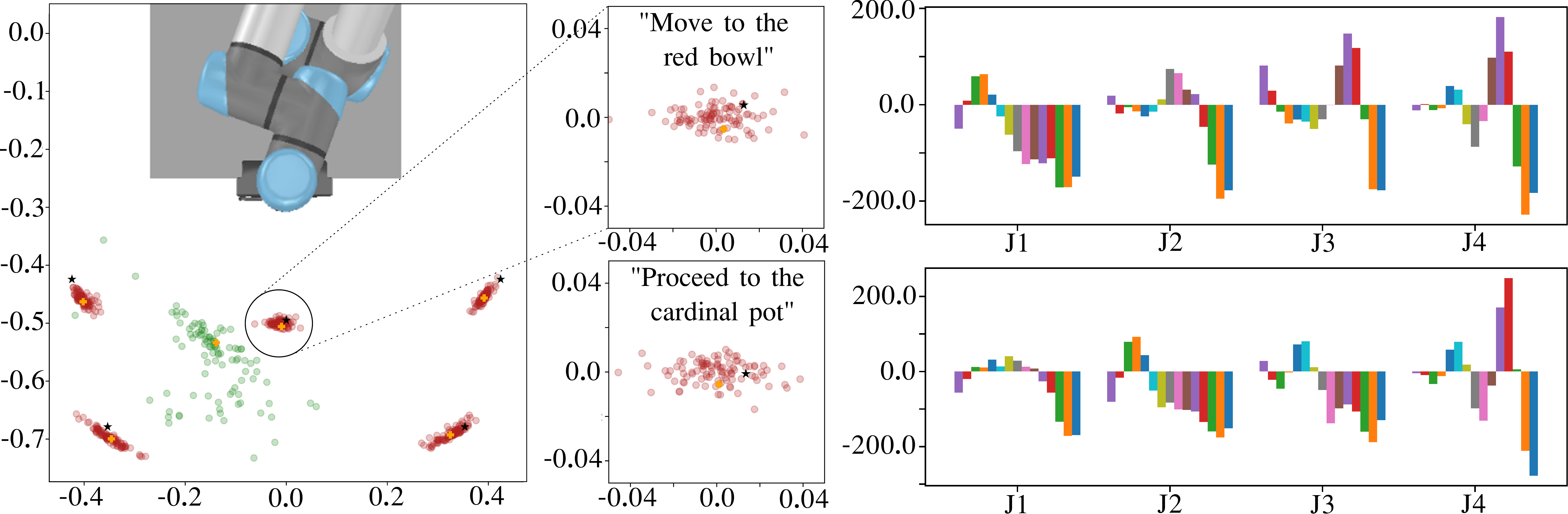}
    \caption{Results for placing an object into bowls at different locations: (Left) Stochastic forward passes allow the model to estimate its certainty about the validity of a task. (Right) Generated weights $\boldsymbol{\Theta}$ for four joints of the DMP shown for two objects close and far away of the robot.}
    \label{fig:results}
\end{figure}

We evaluate our model in a simulated binning task in which the robot is tasked to place a cube into a bowl as outlined by the verbal command. 
Each environment contains between three and five objects differentiated by their size (small, large), shape (round, square) and color (red, green, blue, yellow, pink), totalling in 20 different objects. 
Depending on the generated scenario, combinations of these three features are necessary to distinguish the targets from each other, allowing for tasks of varying complexity. 

To train our model, we generated a dataset of 20,000 demonstrated 7 DOF trajectories (6 robot joints and 1 gripper dimension) in our simulated environment together with a sentence generator capable of creating natural task descriptions for each scenario. In order to create the language generator, we conducted an human-subject study to collect sentence templates of a placement task as well as common words and synonyms for each of the used features. By utilising these data, we are able to generate over 180,000 unique sentences, depending on the generated scenario. 

The generated parameters of the low-level DMP controller -- the weights and goal position -- must be sufficiently accurate in order to successfully deliver the object to the specified bin. On the right side of Figure~\ref{fig:results}, the generated weights for the DMP are shown for two tasks in which the target is close and far away from the robot, located at different sides of the table, indicating the robots ability to generate differently shaped trajectories.
The accuracy of the goal position can be seen in Figure \ref{fig:results}(left) which shows another aspect of our approach: By using stochastic forward passes \citep{gal2015dropout} the model can return an estimate for the validity of a requested task in addition to the predicted goal configuration.
The figure shows that the goal position of a red bowl has a relatively small distribution independently of the used sentence or location on the table, where as an invalid target (green) produces a significantly larger distribution, indicating that the requested task may be invalid. 

To test our model, we generated 500 new scenario testing each of the three features to identify the correct target among other bowls. A task is considered to be successfully completed when the cube is withing the boundaries of the targeted bowl. Bowls have a bounding box of 12.5 and 17.5cm edge length for the small and large variant, respectively.
Our experiments showed that using the objects color or shape to uniquely identify an object allows the robot successfully complete the binning task in 97.6\% and 96.0\% of the cases. However, using the shape alone as a unique identifier, the task could only be completed in 79.0\% of the cases. We suspect that the loss of accuracy is due to the low image resolution of the input image, preventing the network from reliably distinguishing the object shapes.  
In general, our approach is able to actuate the robot with an target error well below 5cm, given the target was correctly identified.

\section{Conclusion and Future Work}
In this work, we presented an imitation learning approach combining language, vision, and motion. A neural network architecture called Multimodal Policy Network was introduced which is able to learn the cross-modal relationships in the training data and achieve high generalization and disambiguation performance as a result. Our experiments showed that the model is able to generalize towards different locations and sentences while maintaining a high success rate of delivering an object to a desired bowl. In addition, we discussed an extensions of the method that allow us to obtain uncertainty information from the model by utilizing stochastic network outputs to get a distribution over the belief.

The modularity of our architecture allows us to easily exchange parts of the network. This can be utilized for transfer learning between different tasks in the semantic network or transfer between different robots by transferring the policy translation network to different robots in simulation, or to bridge the gap between simulation and reality.

\bibliographystyle{plainnat}
\bibliography{references}

\end{document}

%% file: math_commands.tex
\usepackage{amsmath,amsfonts,bm}

\def\eqref#1{equation~\ref{#1}}
\def\1{\bm{1}}

\def\vb{{\bm{b}}}

\def\ve{{\bm{e}}}

\def\vg{{\bm{g}}}

\def\vs{{\bm{s}}}

\def\mI{{\bm{I}}}

\def\mW{{\bm{W}}}

\DeclareMathAlphabet{\mathsfit}{\encodingdefault}{\sfdefault}{m}{sl}
\SetMathAlphabet{\mathsfit}{bold}{\encodingdefault}{\sfdefault}{bx}{n}